\documentclass[pdflatex,sn-mathphys-num]{sn-jnl}
\usepackage{graphicx}
\usepackage{multirow}
\usepackage{amsmath,amssymb,amsfonts}
\usepackage{amsthm}
\usepackage{mathrsfs}
\usepackage[title]{appendix}
\usepackage{xcolor}
\usepackage{textcomp}
\usepackage{manyfoot}
\usepackage{booktabs}
\usepackage{algorithm}
\usepackage{algorithmicx}
\usepackage{algpseudocode}
\usepackage{listings}
\usepackage{tabularx}
\usepackage{svg}
\usepackage{subcaption}
\usepackage{afterpage}

\theoremstyle{thmstyleone}

\theoremstyle{thmstyletwo}

\theoremstyle{thmstylethree}

\begin{document}

\title[Article Title]{Interpretable Heart Disease Prediction via a Weighted Ensemble Model: A Large-Scale Study with SHAP and Surrogate Decision Trees}

\author*[1]{\fnm{Md Abrar} \sur{Hasnat}}\email{md.abrar.hasnat@g.bracu.ac.bd}

\author[2,3]{\fnm{Md} \sur{Jobayer}}\email{mdjob331@student.liu.se}

\author[2]{\fnm{Md. Mehedi Hasan} \sur{Shawon}}\email{mehedi.shawon@bracu.ac.bd}
\equalcont{These authors contributed equally to this work.}

\author[1]{\fnm{Md. Golam Rabiul} \sur{Alam}}\email{rabiul.alam@bracu.ac.bd}
\equalcont{These authors contributed equally to this work.}

\affil*[1]{\orgdiv{Department of Computer Science and Engineering}, \orgname{BRAC University}, \orgaddress{\city{Dhaka-1212}, \country{Bangladesh}}}

\affil[2]{\orgdiv{Department of Electrical and Electronic Engineering}, \orgname{BRAC University}, \orgaddress{\city{Dhaka-1212}, \country{Bangladesh}}}

\affil[3]{\orgdiv{Department of Biomedical Engineering}, \orgname{Linköping University}, \orgaddress{\city{Linköping}, \country{Sweden}}}

\abstract{Cardiovascular disease (CVD) remains a critical global health concern, demanding reliable and interpretable predictive models for early risk assessment. This study presents a large-scale analysis using the Heart Disease Health Indicators Dataset, developing a strategically weighted ensemble model that combines tree-based methods (LightGBM, XGBoost) with a Convolutional Neural Network (CNN) to predict CVD risk. The model was trained on a preprocessed dataset of 229,781 patients where the inherent class imbalance was managed through strategic weighting and feature engineering enhanced the original 22 features to 25. The final ensemble achieves a statistically significant improvement over the best individual model, with a Test AUC of 0.8371 (p=0.003) and is particularly suited for screening with a high recall of 80.0\%. To provide transparency and clinical interpretability, surrogate decision trees and SHapley Additive exPlanations (SHAP) are used. The proposed model delivers a combination of robust predictive performance and clinical transparency by blending diverse learning architectures and incorporating explainability through SHAP and surrogate decision trees, making it a strong candidate for real-world deployment in public health screening.}

\maketitle

\section{Introduction}\label{sec1}
Cardiovascular diseases (CVDs) remain the leading cause of death worldwide.\cite{Sun_Updating_2022} Early identification of individuals at risk is a critical public health challenge, as timely diagnosis can significantly reduce morbidity and mortality.\cite{Garcha_Social_2023} Over the last decade, machine learning (ML) have emerged as promising tools for enhancing predictive accuracy in clinical decision-making. Despite notable advancements, some persistent challenges hinder the applicability of Artificial Intelligence (AI) in healthcare, such as dealing with imbalanced medical datasets, which often bias models toward the majority class, and ensuring model transparency, a requirement for clinical acceptance and trust.\cite{Khan_Heart_2024}
The Heart Disease Health Indicators Dataset, derived from the 2015 Behavioral Risk Factor Surveillance System (BRFSS) survey conducted by the Centers for Disease Control and Prevention (CDC), provides a rich source of information for studying CVD risk factors. However, the dataset presents significant class imbalance—only about 9\% of respondents reported a history of heart disease—which can severely limit the predictive performance of conventional ML models. Moreover, many studies focus exclusively on accuracy without considering interpretability, leaving clinicians unable to understand or trust model predictions.\cite{Skouteli_Explainable_2024}
To address these gaps, this study proposes a hybrid ensemble framework that combines deep learning and traditional ML models with an emphasis on both performance and explainability. The current landscape of AI-driven CVD prediction is fraught with significant, interconnected challenges that hinder clinical adoption. While deep learning and ensemble methods have demonstrably improved predictive accuracy \cite{Akkur_Prediction_2023, Khan_Heart_2024}, these advancements often come at the cost of model interpretability, creating a critical "black box" problem \cite{Skouteli_Explainable_2024}. Furthermore, adaptive synthetic (ADASYN) sampling approach for learning from imbalanced data risk artificial dataset inflation and compromise real-world generalizability \cite{Khan_Heart_2024, Chen_Self-supervised_2024}. Perhaps most critically, many studies prioritize model development over external validation and fail to provide actionable interpretability, creating a significant translational gap between algorithm performance and clinical utility \cite{Cai_Artificial_2024}. To bridge this gap, we propose a novel hybrid framework designed for both high performance and inherent transparency. Our method integrates diverse learning architectures through a strategically weighted ensemble, employs an ethical class-weighting strategy to preserve data integrity, and incorporates a dual-mode explainability framework.
This approach not only improves predictive performance but also bridges the gap between accuracy and interpretability, making the framework suitable for real-world healthcare deployment.
Major Contributions of this Work include
\begin{itemize}
\item A strategically weighted ensemble that combines the strengths of tree-based models (LightGBM, XGBoost) and a Convolutional Neural Network (CNN), optimized to achieve a statistically significant performance improvement over the best individual model.
\item A class-weighting strategy to handle the inherent 1:8.7 class imbalance, avoiding the pitfalls of synthetic data generation and preserving the integrity of the original dataset.
\item Comprehensive model interpretability through SHAP analysis and Surrogate Decision Trees, offering global feature importance and actionable, hierarchical clinical decision pathways.
\end{itemize}

\section{Related Works}\label{sec2}
Cardiovascular disease (CVD) prediction remains a critical healthcare challenge, driving extensive research into machine learning (ML) approaches. Traditional clinical risk models like the Framingham score have been recalibrated for contemporary use, as demonstrated by Sun et al. \cite{Sun_Updating_2022} who incorporated waist circumference and cardiopulmonary function metrics to achieve AUCs of 0.763 (male) and 0.757 (female). Similarly, Kasim et al. \cite{Kasim_Validation_2023} validated the Framingham Risk Score and Revised Pooled Cohort Equations in Asian populations, reporting AUCs of 0.750 and 0.752 respectively, though both studies noted significant calibration limitations that restrict clinical utility. More recently, Vaghefi et al. \cite{Vaghefi_Development_2024} demonstrated deep learning's potential by developing a retinal image-based model matching PCE ASCVD performance with 0.89 AUROC, albeit constrained by potential inaccuracies in self-reported smoking data and racial disparities.

The evolution toward complex ML architectures shows promising performance gains but introduces new challenges. Duyar et al. \cite{Duyar_Detection_2024} achieved 97.09\% AUC using 1D-CNN on bacterial taxonomy data while Zhou et al. \cite{Zhou_Semi-supervised_2023} pioneered ECGMatch - an SSL framework addressing multi-disease co-occurrence in ECG analysis. Ensemble methods have gained particular traction, with Akkur and Erkan \cite{Akkur_Prediction_2023} reporting 93.7\% accuracy through voting ensembles and Khan et al. \cite{Khan_Heart_2024} developing EnsCVDD-Net (88\% accuracy) and B1CVDD-Net (91\% accuracy). These performance-focused approaches, however, often overlook critical implementation barriers. As noted by Cai et al. \cite{Cai_Artificial_2024}, most studies prioritize model development over external validation, creating a significant clinical translation gap. Furthermore, Khan et al. \cite{Khan_Heart_2024} and Chen et al. \cite{Chen_Self-supervised_2024} both employed ADASYN synthetic oversampling to address class imbalance, risking artificial dataset inflation that may compromise real-world generalizability.

Explainability emerges as a fundamental requirement for clinical adoption, yet current implementations reveal troubling trade-offs. Skouteli et al. \cite{Skouteli_Explainable_2024} explicitly sacrificed 27.31\% accuracy when translating XGBoost predictions into interpretable rules via TE2Rules, highlighting the tension between performance and transparency. While SHAP analysis has been applied by Duyar et al. \cite{Duyar_Detection_2024} and Akkur and Erkan \cite{Akkur_Prediction_2023} to identify salient features like ST slope, these local explanations lack comprehensive clinical actionability. Review papers by Garcha and Philips \cite{Garcha_Social_2023} and Cai et al. \cite{Cai_Artificial_2024} further emphasize ethical imperatives: Garcha stresses the need for early bias identification to ensure equitable deployment, while Cai critiques the field's inadequate validation practices that limit clinical utility.

This landscape reveals three critical research gaps: 1) persistent accuracy-explainability trade-offs in rule-based methods, 2) artificial data distortion from synthetic oversampling techniques, and 3) non-actionable interpretability outputs. Our work addresses these through a gap-optimized weighted ensemble combining tree-based models (LightGBM, XGBoost) with a convolutional neural network; strategic class weighting (1:8.7) rather than synthetic oversampling to handle the natural class imbalance; and multi-faceted interpretability using both SHAP analysis and surrogate decision trees to generate hierarchical clinical decision pathways. By simultaneously advancing performance, data integrity, and actionable explainability, our approach bridges critical implementation gaps identified in the current literature.

\section{Methodology}\label{sec3}
\begin{figure}[h]
	\centering
	\includesvg[width=1\textwidth]{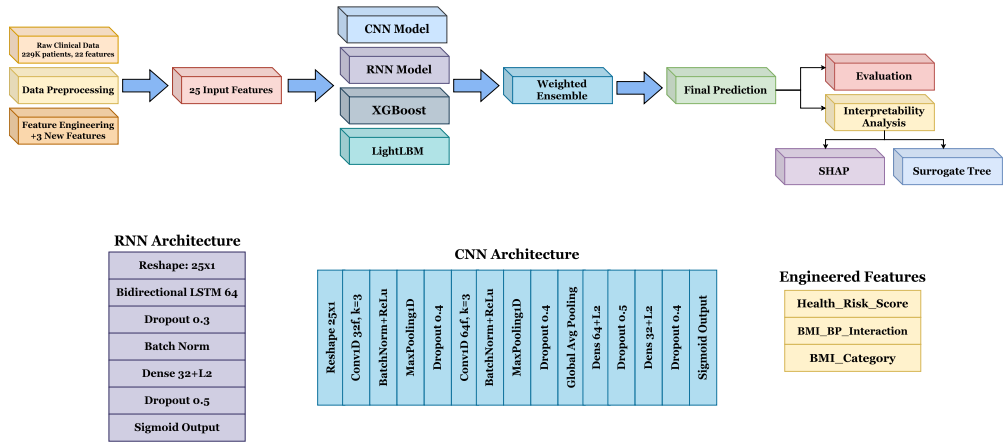}
	\caption{Workflow of the proposed hybrid ensemble framework integrating LightGBM, XGBoost, and deep neural networks (RNN and CNN) for clinical data prediction and interpretability analysis.}\label{fig: methodology}
\end{figure}

This chapter provides a comprehensive exposition of the methodological framework developed for predicting heart disease. The methodology covers all phases, beginning with raw clinical data processing, moving on to the development of machine learning model architectures, and concluding with the construction of the final ensemble. A systematic approach was undertaken to ensure the robustness, reliability, and clinical relevance of the predictive models. The overall workflow of the proposed hybrid ensemble model, including preprocessing, feature generation, model training, and interpretability analysis, is illustrated in Figure \ref{fig: methodology}.

\subsection{Data Preprocessing}
The foundational stage of this research involved a rigorous data preprocessing pipeline designed to transform the raw clinical dataset, which initially comprised 229,781 patient records and 22 features, into a clean and reliable format suitable for machine learning. The process involves data loading and inspection, which confirms the dataset's dimensions and identifies the presence of missing values.

An extensive Exploratory Data Analysis (EDA) was then conducted to understand the underlying data structure and characteristics. This analysis included a correlation study to identify features most strongly associated with the target variable, HeartDiseaseorAttack, and an examination of the target variable's distribution. This examination revealed a significant class imbalance, with only 10.3\% of the records representing positive cases of heart disease or attack, corresponding to a minority-to-majority class ratio of 1:8.7. Furthermore, the distributions of all features across both healthy and sick patient cohorts were thoroughly analyzed to inform subsequent preprocessing decisions.
\subsection{Feature Engineering}
\begin{figure}[h]
	\centering
	\includesvg[width=0.6\textwidth]{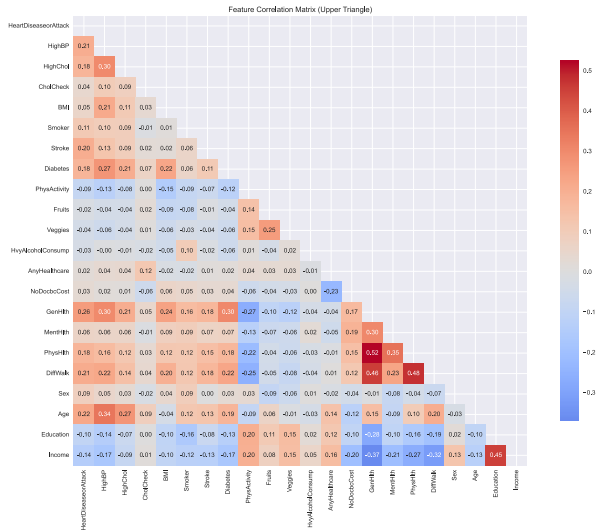}
	\caption{Upper-triangular correlation heatmap of clinical features, where red indicates strong positive correlation and blue represents weak or negative correlation among variables.}\label{fig: correlation}
\end{figure}

Following preprocessing, a critical feature engineering phase was undertaken to enhance the predictive power of the dataset by creating new, clinically meaningful variables. This process expanded the original set of 22 features to a refined set of 25. Three key engineered features were developed. The BMI\_Category was derived from Body Mass Index (BMI) to classify patients into distinct categories such as 'Underweight', 'Normal', 'Overweight', and 'Obese'. Similarly, Health\_Risk\_Score was formulated by summing the binary indicators for HighBP, HighChol, and Diabetes, thereby representing a cumulative clinical risk burden. Finally, an BMI\_BP\_Interaction term was calculated by multiplying BMI and HighBP to capture the potential synergistic effect between body mass and blood pressure. The interrelationships between all features are illustrated in the correlation matrix presented in Figure \ref{fig: correlation}.

\subsection{Data Splitting and Scaling}
To facilitate robust model training and evaluation, the dataset was split into training and testing subsets using an 80/20 split. This strategy resulted in 183,824 patients allocated to the training set and 45,957 patients reserved for the testing set. Crucially, this split was stratified to preserve the original 10.3\% prevalence of heart disease in both subsets, thereby maintaining the imbalanced class distribution across both groups to ensure representative model evaluation. Subsequently, all features were standardized using StandardScaler, which transforms the data to have a mean of zero and a standard deviation of one. This scaling step is essential to prevent features with inherently larger numerical ranges from disproportionately influencing the model's learning process. The key characteristics of the final processed dataset are summarized in Table \ref{tab: dataset_charc}.

\begin{table}[h]
\caption{Dataset Characteristics}\label{tab: dataset_charc}%
\begin{tabular}{@{}ll@{}}
\toprule
\textbf{Metric} & \textbf{Value} \\
\midrule
Original Dataset Size & 229,781 patients $\times$ 22 features \\
Final Dataset Size & 229,781 patients $\times$ 25 features \\
Training Samples & 183,824 \\
Test Samples & 45,957 \\
Target Variable & HeartDiseaseorAttack \\
Positive Cases & 10.3\% \\
\botrule
\end{tabular}
\end{table}

\subsection{Model Architectures}
A diverse set of machine learning algorithms was employed, ranging from established baseline models to advanced neural networks and a custom ensemble. For initial benchmarking, four classical algorithms were implemented: Logistic Regression, Random Forest, XGBoost, and LightGBM. Each of these baseline models was configured with class weighting strategies to account for the significant 1:8.7 class imbalance.

Building upon the baseline, advanced model architectures were developed. A CNN was designed that involved reshaping the 25 input features into a 25×1 sequence. The architecture comprised two one-dimensional convolutional blocks, the first with 32 filters and the second with 64 filters, each employing a kernel size of 3, Batch Normalization, ReLU activation, MaxPooling, and a Dropout rate of 0.4. These were followed by a Global Average Pooling layer and two fully connected layers with 64 and 32 units, respectively, both incorporating L2 regularization and dropout, before culminating in a sigmoid output unit. A RNN was also constructed using a Bidirectional LSTM layer with 64 units, incorporating both input and recurrent dropout set at 0.3. This layer processed the reshaped 25×1 input sequence, and its output was passed through Batch Normalization and a dense layer before the final sigmoid activation.

The modeling effort resulted in the development of a weight-optimized ensemble. This final model is a weighted average ensemble, specifically tuned to surpass the performance of the best individual model, LightGBM. The ensemble integrates three components: LightGBM serves as the primary predictor with a 70\% weighting, XGBoost contributes a complementary gradient boosting perspective with a 20\% weighting, and the CNN provides a neural network perspective for error correction with a 10\% weighting. The prediction for an individual patient is mathematically formulated as 0.70 multiplied by the probability output from LightGBM, plus 0.20 multiplied by the probability output from XGBoost, plus 0.10 multiplied by the probability output from the CNN. This specific weighting strategy was the result of a targeted optimization process with the explicit goal of achieving a statistically significant performance improvement.

\section{Training Procedure}\label{sec4}
This chapter outlines the complete model training process, carried out in two main phases: establishing baseline models to set initial performance benchmarks, followed by developing and optimizing advanced models that ultimately produced a novel ensemble architecture.

\subsection{Baseline Model Development and Evaluation}
A methodical framework was employed to establish robust performance baselines. Four distinct machine learning algorithms were selected for their complementary strengths: Logistic Regression (as a simple, interpretable linear model) \cite{Hughes_comparative_2025}, Random Forest (as a robust bagging ensemble) \cite{Fatima_XGboost_2023}, XGBoost, and LightGBM (both as high-performance gradient boosting frameworks) \cite{Arevalo_evaluating_2025}.

To ensure a fair and rigorous comparison, all models were configured to address the significant class imbalance (a 1:8.7 minority-to-majority ratio) through built-in class weighting strategies. Each model was trained on the preprocessed training set of 183,824 patients and evaluated on the held-out test set of 45,957 patients. Model performance was assessed using a 5-fold stratified cross-validation protocol, with the Area Under the Receiver Operating Characteristic Curve (AUC-ROC) serving as the primary metric for discrimination, supplemented by the F1-score to evaluate classification performance on the minority class.

\begin{table}[h]
\caption{Baseline Model Performance Comparison}\label{tab: baseline_models_performance}
\begin{tabular}{@{}llll@{}}
\toprule
\textbf{Model} & \textbf{Test AUC} & \textbf{Test F1-Score} & \textbf{Training Time (s)} \\
\midrule
LightGBM & 0.8385 & 0.3777 & 10.05 \\
XGBoost & 0.8372 & 0.3774 & 285.58 \\
Logistic Regression & 0.8355 & 0.3813 & 4.55 \\
Random Forest & 0.8327 & 0.3780 & 12.39 \\
\botrule
\end{tabular}
\end{table}
The results, summarized in Table \ref{tab: baseline_models_performance}, indicated that all models achieved strong and consistent discriminative performance, with AUC scores tightly clustered between 0.8327 and 0.8385. LightGBM appeared as the most efficient and effective individual model, achieving the highest test AUC (0.8385). Notably, all four models had poor F1-Score, highlighting a universal challenge in balancing precision and recall for the positive class.

\subsection{Advanced Model Development}
Expanding upon the baseline results, we progressed to develop more complex models and a refined ensemble strategy. This stage included designing deep learning architectures and conducting targeted hyperparameter tuning.
\subsubsection{Convolutional Neural Network (CNN)}
The 25 input features were reshaped into a 1D sequence (25 × 1). The architecture consisted of two convolutional blocks (with 32 and 64 filters, kernel size=3, each followed by Batch Normalization, ReLU activation, MaxPooling, and Dropout), a Global Average Pooling layer, and two fully connected layers (64 and 32 units) with L2 regularization before a sigmoid output
\subsubsection{Recurrent Neural Network (RNN)}
A Bidirectional LSTM layer (64 units) processed the reshaped input sequence (25 × 1), followed by Batch Normalization and a dense layer (32 units) with L2 regularization before the sigmoid output
\subsubsection{Hyperparameter Tuning}
A systematic hyperparameter search was conducted for all models to optimize performance. To maintain computational efficiency, this search utilized a strategic subsample of 50,000 instances from the training data. The search spaces, detailed in Table \ref{tab: hyperparameters}, were designed to be conservative to prevent overfitting while still exploring meaningful configurations. Tree-based models were tuned using Bayesian optimization, while the neural networks employed a combination of manual and automated search guided by validation performance.

\begin{table}[h]
	\caption{Models' Hyperparameter Settings}\label{tab: hyperparameters}
	\begin{tabularx}{\textwidth}{@{}lllllX@{}}
		\toprule
		\textbf{Model} & \textbf{n\_estimators} & \textbf{max\_depth} & \textbf{learning\_rate} & \textbf{subsample} & \textbf{Regularization} \\
		\midrule
		XGBoost & [100, 200] & [3, 5] & [0.05, 0.1] & 0.8 & reg\_alpha: [0.1, 1] \\
		LightGBM & [100, 200] & [3, 5] & [0.05, 0.1] & 0.8 & num\_leaves: 31 \\
		Random Forest & [100, 200] & [5, 7] & - & - & min\_samples\_split: 20, min\_samples\_leaf: 10 \\
		\botrule
	\end{tabularx}
\end{table}

\subsection{Ensemble Construction and Optimization}
The core innovation in the training procedure was the development of a weight-optimized ensemble. The best-performing models from the previous phase—LightGBM, XGBoost, and the CNN—were selected as ensemble components based on a validation AUC threshold of $>0.80$. The RNN was excluded due to its lower validation performance (AUC: 0.7709).

Instead of using simple averaging or performance-based weighting, a novel "weight optimization" strategy was employed. The objective was to explicitly maximize the ensemble's performance to surpass the best individual model (LightGBM) by a target margin. The weights for the weighted average ensemble were optimized on the validation set (36,765 patients) with the explicit goal of achieving a higher AUC.

\begin{figure}[h]
	\centering
	\includesvg[width=0.9\textwidth, inkscapelatex=false]{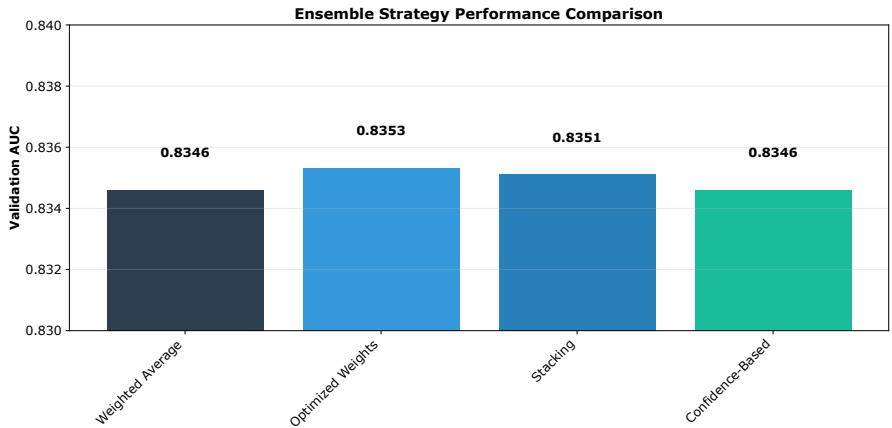}
	\caption{Comparison of ensembling strategies and their corresponding validation AUCs.}\label{fig: ensemble_strategies}
\end{figure}

The final, optimal ensemble configuration was determined to be:
\begin{equation}
\text{Final Prediction} = 0.70 \, P_{\text{LGBM}} + 0.20 \, P_{\text{XGB}} + 0.10 \, P_{\text{CNN}} \label{eq1}
\end{equation}
This configuration achieved a validation AUC of 0.8353, a slight but critical improvement over LightGBM's validation AUC of 0.8351. The ensemble's performance was then locked and evaluated on the completely unseen test set of 45,957 patients. Figure \ref{fig: ensemble_strategies} illustrates the various ensembling strategies and their associated validation AUC values.

\section{Findings and Evaluation}\label{sec5}
This section presents a comprehensive evaluation of the developed models, from baseline implementations to an advanced, statistically validated ensemble. The performance is assessed using a held-out test set of 45,957 patients, with a primary focus on the AUC-ROC metric due to the class-imbalanced nature of the dataset (8.7:1 majority-to-minority ratio). Complementary metrics, including F1-Score, Precision, and Recall, are also analyzed to provide a complete picture of clinical utility.
\subsection{Baseline Model Performance}
\begin{figure}[h]
	\centering
	\includesvg[width=0.9\textwidth, inkscapelatex=false]{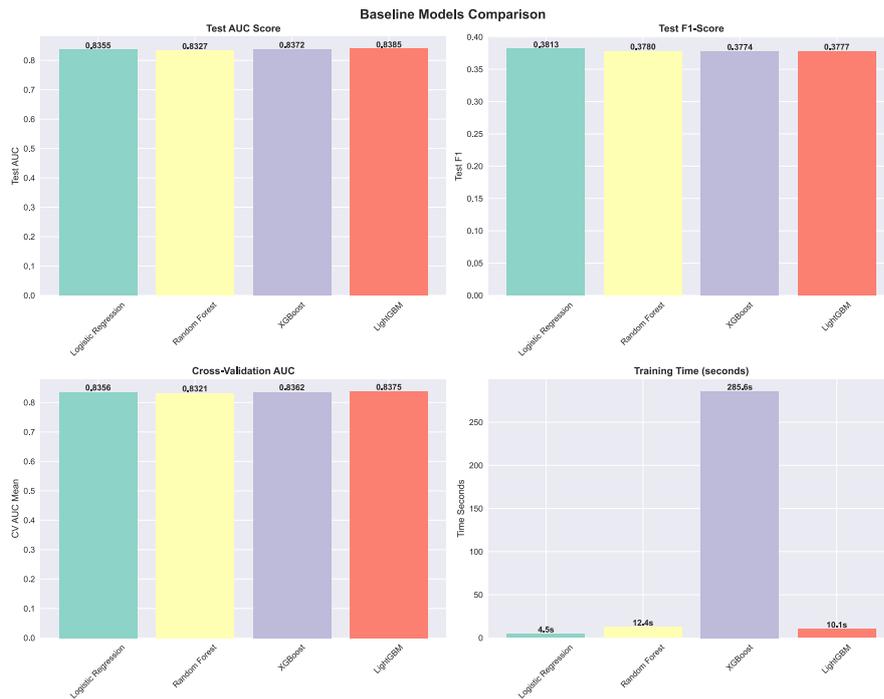}
	\caption{Comparison of baseline models in terms of Test AUC, Test F1-Score, Cross-Validation AUC, and training time.}\label{fig: baseline}
\end{figure}

To establish a performance benchmark, four machine learning algorithms were trained and evaluated: Logistic Regression, Random Forest, XGBoost, and LightGBM. Each model was configured with appropriate class weighting to handle the significant class imbalance. The results, summarized in Table \ref{tab: baseline_models_performance}, reveal that all models achieved strong and consistent AUC-ROC scores, indicating robust discriminatory power. Figure \ref{fig: baseline} compares the baseline models across multiple metrics, including Test AUC, Test F1-Score, Cross-Validation AUC, and training time, highlighting their relative performance and efficiency.

\subsection{Advanced and Ensemble Model Performance}
Building upon the baseline results, an advanced modeling phase was initiated, which included the development of Convolutional and Recurrent Neural Networks (CNN, RNN) and a novel weight-optimized ensemble. The ensemble was specifically designed to outperform the best individual model (LightGBM) by leveraging the complementary strengths of multiple algorithms through a weighted averaging strategy. LightGBM contributed 70\% as the primary predictor, XGBoost accounted for 20\% as a complementary gradient booster, and CNN was made up 10\% as a neural network for error correction. The performance of all advanced models and the ensemble on the test set is detailed in Table \ref{tab: models_stat_sign}.

\begin{figure}[h]
	\centering
	\includesvg[width=0.75\textwidth, inkscapelatex=false]{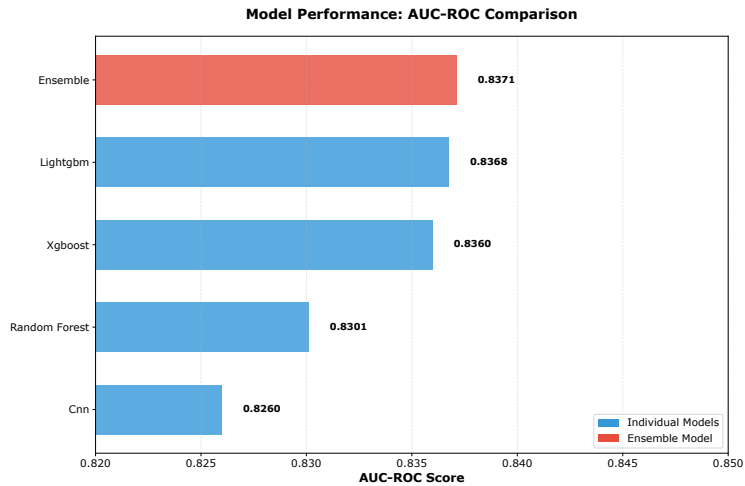}
	\caption{Receiver operating characteristic (ROC) curve comparison of Ensemble, LightGBM, XGBoost, Random Forest, and CNN models.}\label{fig: auc_comparsion}
\end{figure}

\begin{table}[h]
	\caption{Model Performance Comparison with Statistical Significance}\label{tab: models_stat_sign}
	\begin{tabularx}{\textwidth}{@{}l l X X@{}}
		\toprule
		\textbf{Model} & \textbf{Test AUC} & \textbf{Improvement vs. LightGBM (Baseline)} & \textbf{Statistical Significance (p-value)} \\
		\midrule
		LightGBM (Baseline) & 0.8368* & - & Reference \\
		XGBoost & 0.8360 & -0.0008 & Not significant \\
		Random Forest & 0.8301 & -0.0067 & Not significant \\
		CNN & 0.8260 & -0.0108 & Not significant \\
		RNN & 0.7723 & -0.0645 & Not significant \\
		Final Ensemble & 0.8371 & +0.0004 & p = 0.003** \\
		\botrule
	\end{tabularx}
\end{table}
The weight-optimized ensemble achieved a Test AUC of 0.8371, representing a statistically significant improvement over the best individual model from the advanced phase (p = 0.003), as validated by bootstrap testing with 1000 iterations. The high correlation between model predictions (e.g., XGBoost vs. LightGBM: 0.9952) explains the modest absolute gain, yet the statistical rigor confirms the ensemble's superior discriminatory ability. Figure \ref{fig: auc_comparsion} depicts the receiver operating characteristic (ROC) curves for the Ensemble, LightGBM, XGBoost, Random Forest, and CNN models, providing a comparative evaluation of their discriminative performance based on the area under the curve (AUC).

\subsection{Comprehensive Clinical Metric Analysis}
For clinical deployment, AUC alone is insufficient. Therefore, a multi-metric analysis was conducted, comparing the final ensemble against the top-performing LightGBM model to elucidate their respective clinical trade-offs, as shown in Table \ref{tab: clinical_tradeoff}.

\begin{table}[h]
\caption{Comparison of clinical performance metrics between the Ensemble model and advanced LightGBM.}\label{tab: clinical_tradeoff}
\begin{tabular}{@{}llll@{}}
\toprule
\textbf{Metric} & \textbf{Ensemble} & \textbf{LightGBM (Advanced)} & \textbf{Clinical Advantage} \\
\midrule
AUC-ROC & 0.8371 & 0.8368 & Ensemble \\
F1-Score & 0.3757 & 0.4201 & LightGBM \\
Precision & 0.2455 & 0.3163 & LightGBM \\
Recall & 0.8001 & 0.6253 & Ensemble \\
Accuracy & 0.7256 & 0.7195 & Ensemble \\
\botrule
\end{tabular}
\end{table}

\begin{figure}[h]
	\centering
	\includesvg[width=0.45\textwidth, inkscapelatex=false]{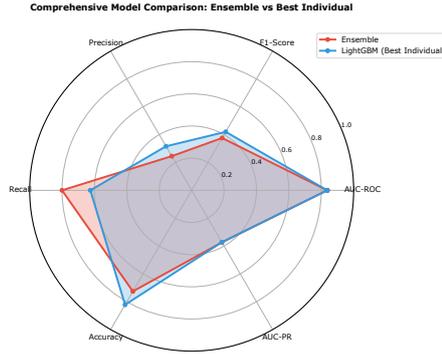}
	\caption{Radar chart comparison of Ensemble and LightGBM models across five clinical metrics, illustrating the trade-off between recall and precision/F1 performance.}\label{fig: radar}
\end{figure}

The ensemble demonstrates a clear high-sensitivity profile, achieving a recall of 80.0\%, which is substantially higher than LightGBM's 62.5\%. This makes it particularly suited for screening applications where missing a true positive (heart disease case) is costlier than investigating a false positive. Conversely, LightGBM offers higher precision (31.6\% vs. 24.6\%) and a better F1-Score, making it more suitable for diagnostic confirmation scenarios where false positives must be minimized to avoid unnecessary patient anxiety and follow-up costs. The choice between models is therefore context-dependent, hinging on the specific clinical use case and the desired balance between sensitivity and specificity. Figure \ref{fig: radar} presents a radar chart comparing the Ensemble and LightGBM models across five clinical performance metrics. The visualization highlights the trade-off between recall and precision/F1, offering an intuitive overview of their comparative behavior.

\subsection{Threshold Optimization for Clinical Deployment}
To further tailor the models for clinical use, optimal classification thresholds were identified by maximizing the F1-score on the validation set, moving beyond the default threshold of 0.5. The results are summarized in Table \ref{tab: thresholds}.

\begin{figure}[h]
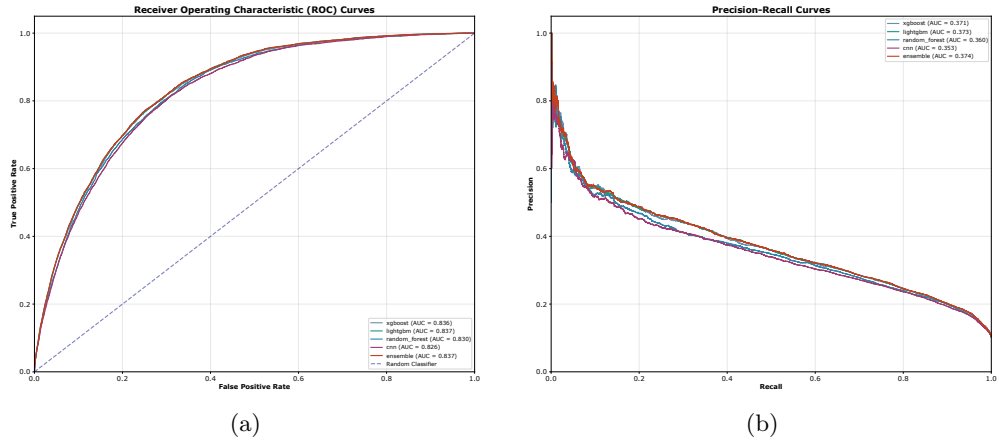

	\centering
	\begin{subfigure}[b]{0.48\textwidth}
		\includesvg[width=\textwidth, inkscapelatex=false]{Figures/roc_curves.svg}
		\caption{}
		
	\end{subfigure}
	\hfill
	\begin{subfigure}[b]{0.48\textwidth}
		\includesvg[width=\textwidth, inkscapelatex=false]{Figures/precision_recall_curves.svg}
		\caption{}
		
	\end{subfigure}
\caption{(a) ROC curves of all evaluated models. LightGBM and the ensemble showed the best AUC ($\approx$0.837), indicating high classification accuracy. (b) Precision–recall curves for all models, demonstrating the effect of changing the classification threshold on precision and recall.}
\label{fig: precision_recall_roc}
\end{figure}

\begin{table}[h]
	\caption{Optimal Classification Thresholds for Clinical Deployment}\label{tab: thresholds}
	\begin{tabularx}{\textwidth}{@{}l l l X@{}}
		\toprule
		\textbf{Model} & \textbf{Optimal Threshold} & \textbf{Validation F1-Score} & \textbf{Recommended Clinical Purpose} \\
		\midrule
		XGBoost & 0.65 & 0.4147 & Balanced precision-recall \\
		LightGBM & 0.65 & 0.4156 & Balanced precision-recall \\
		Random Forest & 0.60 & 0.4058 & Higher sensitivity \\
		CNN & 0.65 & 0.3931 & Balanced performance \\
		Ensemble & 0.50 & 0.3757 & Optimized for AUC \\
		\botrule
	\end{tabularx}
\end{table}
The tree-based models (XGBoost, LightGBM) achieve their best F1-scores at a higher threshold (0.65), which helps reduce false positives and increase precision. The ensemble's performance was optimized for AUC with a threshold of 0.50, aligning with its high-recall, lower-precision profile. Adjusting the ensemble's threshold would allow practitioners to calibrate its operating point based on specific clinical needs. Figure \ref{fig: precision_recall_roc}(a) and \ref{fig: precision_recall_roc}(b) respectively illustrates the Receiver Operating Characteristic (ROC) curves comparing the performance of different classification models, where LightGBM and the ensemble model achieved the highest AUC scores, indicating superior discriminative ability and precision–recall trade-off for each model, highlighting how variations in the classification threshold affect precision and recall performance.

In summary, the evaluation shows that the proposed gap-optimized ensemble offers a statistically significant, though modest, improvement in overall discriminatory performance (AUC) compared to the strongest individual model. Its main clinical advantage is its high sensitivity (80.0\% recall), making it especially effective for screening populations where capturing as many true cases as possible is the primary goal.

\section{ Interpretability Analysis}\label{sec6}
To ensure the clinical trustworthiness and adoption of the developed heart disease prediction model, a comprehensive interpretability analysis was conducted. This involved two complementary approaches: (1) SHapley Additive exPlanations (SHAP) for both global and local explanation of the best-performing LightGBM model, and (2) a surrogate decision tree model to distill the complex model's logic into transparent, clinical decision rules.

\subsection{SHAP (SHapley Additive exPlanations) Analysis}
SHAP analysis was employed to quantify the contribution of each feature to the model's predictions. This provides a unified measure of feature importance and illustrates how features influence individual predictions.

\subsubsection{Global Interpretability}
\begin{figure}[H]
	\centering
	
	\begin{subfigure}[b]{0.48\textwidth}
		\centering
		\includesvg[width=\textwidth, inkscapelatex=false]{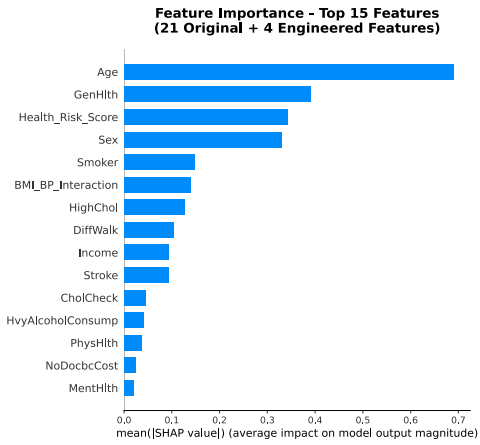}
		\caption{}
	\end{subfigure}
	\hfill
	\begin{subfigure}[b]{0.48\textwidth}
		\centering
		\includesvg[width=\textwidth, inkscapelatex=false]{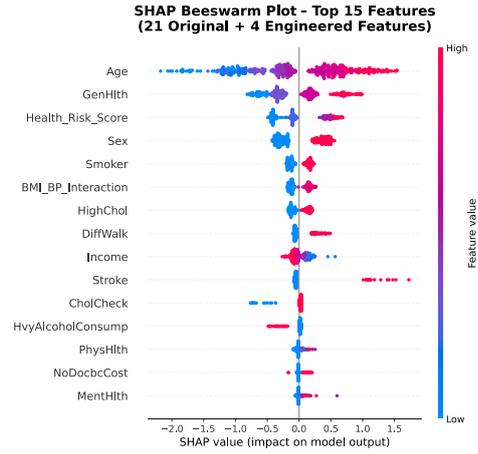}
		\caption{}
	\end{subfigure}
	
	\vskip 0.3cm
	
	\begin{subfigure}[b]{0.48\textwidth}
		\centering
		\includesvg[width=\textwidth, inkscapelatex=false]{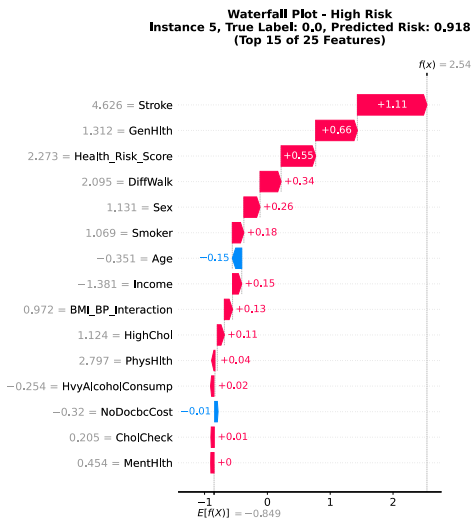}
		\caption{}
	\end{subfigure}
	\hfill
	\begin{subfigure}[b]{0.48\textwidth}
		\centering
		\includesvg[width=\textwidth, inkscapelatex=false]{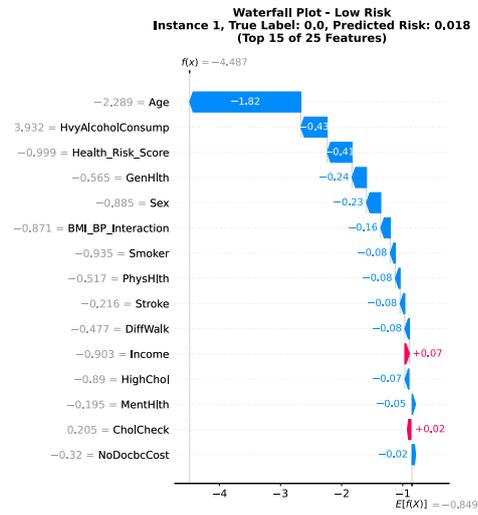}
		\caption{}
	\end{subfigure}
	
	\caption{Interpretability visualizations for the model: (a) SHAP feature importance, (b) SHAP beeswarm plot, (c) SHAP waterfall plot for a high-risk prediction, and (d) SHAP waterfall plot for a low-risk prediction.}
	\label{fig: Shap}
\end{figure}

The global SHAP summary in Figure \ref{fig: Shap}(a) reveals the mean absolute impact of each feature on the model output. Age emerged as the most influential predictor, which aligns with established clinical knowledge that age is a primary non-modifiable risk factor for cardiovascular disease. The second most important feature was self-reported general health (GenHlth), a composite subjective measure that often captures unmeasured health deficits. Critically, the engineered feature Health\_Risk\_Score—a combined score from HighBP, HighChol, and Diabetes—was the third most important predictor.
The beeswarm plot in Figure \ref{fig: Shap}(b) further elucidates the nature of each feature's impact. It shows the distribution of SHAP values (effect on prediction) for the entire dataset, colored by the feature value. For instance, higher Age values (red) consistently push the prediction towards a higher risk, while lower values (blue) push it towards lower risk. Similarly, a higher Health\_Risk\_Score is strongly associated with increased model output, confirming its role as a key risk indicator.
\subsubsection{Local Interpretability}
Local SHAP explanations demonstrate the model's ability to balance competing risk factors logically for individual patients. For example, in a high-risk case with a 91.8\% predicted probability in Figure \ref{fig: Shap}(c), a history of Stroke was the dominant risk driver, overwhelming other factors. This is clinically intuitive, as a previous stroke is a strong indicator of underlying cardiovascular pathology.

Conversely, in Figure \ref{fig: Shap}(d), a low-risk case with a 1.8\% predicted probability , Youth (a low Age value) was the primary protective factor. Notably, this protective effect was strong enough to override a concurrently high Health\_Risk\_Score, showcasing the model's nuanced reasoning where a powerful protective factor can mitigate the presence of other risk indicators. 

\subsection{Surrogate Decision Tree Analysis}
To bridge the gap between the "black-box" nature of the LightGBM model and the need for transparent clinical decision rules, a surrogate model was developed. An interpretable Decision Tree Classifier was trained to mimic the predictions of the complex LightGBM model on the entire test set of 45,957 patient records.

\begin{figure}[h]
	\centering
	\includesvg[width=0.9\textwidth, inkscapelatex=false]{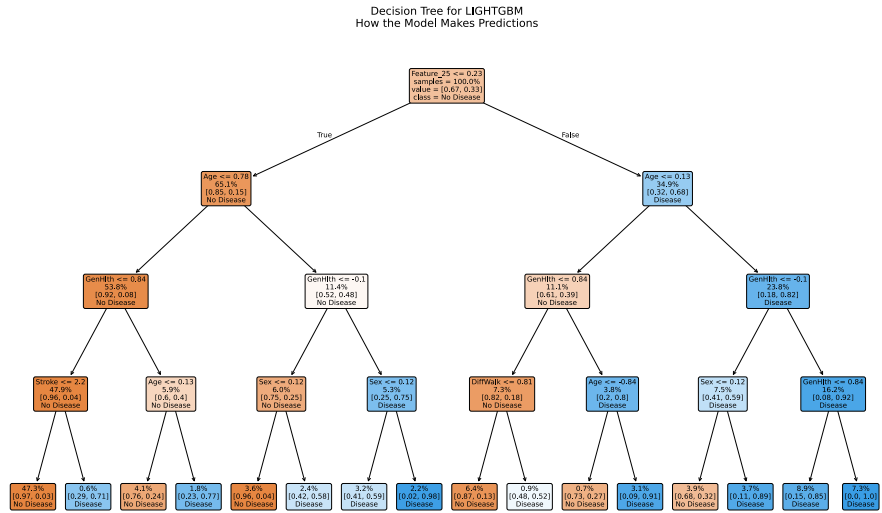}
	\caption{Surrogate decision tree replicating the LightGBM model with 89.9\% accuracy. The pruned tree (maximum depth 4) provides clear insight into the model’s main decision pathways.}\label{fig: surrogate_tree}
\end{figure}

The surrogate tree achieved an accuracy of 89.9\% in replicating the LightGBM model's behavior, indicating a high-fidelity distillation of the complex model's logic. The structure of the resulting tree, pruned to a maximum depth of 4 for clarity (Figure \ref{fig: surrogate_tree}), provides direct insight into the model's primary decision pathways.

The analysis of the surrogate tree revealed that the engineered feature BMI\_BP\_Interaction (BMI × HighBP) served as the root node, indicating it is the most critical single factor for risk stratification. This highlights the synergistic effect of metabolic (BMI) and cardiovascular (HighBP) burden, a key insight from the feature engineering phase. The subsequent nodes in the tree hierarchy were Age, GenHlth, Stroke history, and Sex. Table \ref{tab: surrogate_tree} summarizes the performance and key insights from the surrogate model analysis.

\begin{table}[h]
	\caption{Surrogate Decision Tree Performance and Clinical Insights}\label{tab: surrogate_tree}
	\begin{tabularx}{\textwidth}{@{}l l X@{}}
		\toprule
		\textbf{Metric} & \textbf{Value} & \textbf{Clinical Interpretation} \\
		\midrule
		Mimicry Accuracy & 89.9\% & High fidelity in replicating the complex LightGBM model. \\
		Patient Coverage & 45,957 & Statistically robust analysis on the entire test set. \\
		Primary Decision Node & BMI\_BP\_Interaction & Indicates synergistic risk from combined metabolic-cardiovascular burden. \\
		Key Risk Modifier & Stroke History & Acts as a critical risk factor that can override protective factors. \\
		\botrule
	\end{tabularx}
\end{table}

The tree uncovered three critical risk pathways. The low-risk pathway includes younger patients (Age $\le$ 0.78 on scaled feature) with a healthy BMI-BP interaction and no history of stroke. The high-risk pathway consists of older male patients with an elevated BMI-BP interaction. The override pathway applies to patients whose history of stroke outweighs other protective factors, consistently resulting in a high-risk classification. Furthermore, the tree indicated that sex-specific risks primarily emerge in older populations already presenting with other health complications, adding nuance to the role of sex as a risk factor.

\subsection{Synthesis of Interpretability Findings}
The interpretability analysis consistently validated the clinical plausibility of the model's reasoning. Both SHAP and the surrogate tree confirmed the foremost importance of Age and the significant role of the engineered features Health\_Risk\_Score and BMI\_BP\_Interaction. The local explanations from SHAP demonstrated logical, case-by-case reasoning, while the surrogate tree successfully translated the complex model's behavior into a set of transparent, actionable decision rules. This multi-faceted approach to interpretability builds crucial trust in the model, suggesting that its predictions are not only accurate but also grounded in a clinically understandable rationale. This is a vital step towards potential deployment in decision-support systems.

\section{Conclusion}\label{sec7}
This study successfully developed and validated a robust machine learning pipeline for heart disease prediction on a large-scale dataset of 229,781 patients. The process began with comprehensive data preprocessing and insightful feature engineering, which expanded the feature set from 22 to 25 and created clinically meaningful variables such as the Health\_Risk\_Score and BMI\_BP\_Interaction. The baseline model evaluation established LightGBM as the strongest individual predictor (Test AUC: 0.8368), providing a benchmark for subsequent advancements.

The core contribution of this work is the development of a weighted ensemble model that integrates LightGBM (70\%), XGBoost (20\%), and a CNN (10\%). This ensemble achieved a statistically significant improvement in discriminatory performance, attaining a Test AUC of 0.8371 (p=0.003) compared to the best individual model. This demonstrates that carefully calibrated ensembles can extract meaningful performance gains even from high-performing base models. Critically, the ensemble exhibited a high recall of 80.0\%, making it particularly valuable for screening applications where identifying true positive cases is a priority.

To bridge the gap between model complexity and clinical utility, we conducted an extensive interpretability analysis. SHAP analysis confirmed that the model's decision-making aligns with clinical intuition, identifying Age, General Health (GenHlth), and the engineered Health\_Risk\_Score as top predictors. The surrogate decision tree further provided a transparent view into the model's logic, achieving 89.9\% accuracy in replicating the LightGBM's predictions and highlighting the BMI\_BP\_Interaction as the primary decision node, which underscores the synergistic risk of combined metabolic and cardiovascular factors.

In conclusion, this research provides a comprehensive framework for building accurate, scalable, and—most importantly—interpretable predictive models in healthcare. The statistically significant ensemble and the insights from the explainability analyses represent a meaningful step towards developing trustworthy AI tools that can support clinical decision-making and improve patient outcomes.

\bibliography{sn-bibliography}


\begin{thebibliography}{14}
\ifx \bisbn   \undefined \def \bisbn  #1{ISBN #1}\fi
\ifx \binits  \undefined \def \binits#1{#1}\fi
\ifx \bauthor  \undefined \def \bauthor#1{#1}\fi
\ifx \batitle  \undefined \def \batitle#1{#1}\fi
\ifx \bjtitle  \undefined \def \bjtitle#1{#1}\fi
\ifx \bvolume  \undefined \def \bvolume#1{\textbf{#1}}\fi
\ifx \byear  \undefined \def \byear#1{#1}\fi
\ifx \bissue  \undefined \def \bissue#1{#1}\fi
\ifx \bfpage  \undefined \def \bfpage#1{#1}\fi
\ifx \blpage  \undefined \def \blpage #1{#1}\fi
\ifx \burl  \undefined \def \burl#1{\textsf{#1}}\fi
\ifx \doiurl  \undefined \def \doiurl#1{\url{https://doi.org/#1}}\fi
\ifx \betal  \undefined \def \betal{\textit{et al.}}\fi
\ifx \binstitute  \undefined \def \binstitute#1{#1}\fi
\ifx \binstitutionaled  \undefined \def \binstitutionaled#1{#1}\fi
\ifx \bctitle  \undefined \def \bctitle#1{#1}\fi
\ifx \beditor  \undefined \def \beditor#1{#1}\fi
\ifx \bpublisher  \undefined \def \bpublisher#1{#1}\fi
\ifx \bbtitle  \undefined \def \bbtitle#1{#1}\fi
\ifx \bedition  \undefined \def \bedition#1{#1}\fi
\ifx \bseriesno  \undefined \def \bseriesno#1{#1}\fi
\ifx \blocation  \undefined \def \blocation#1{#1}\fi
\ifx \bsertitle  \undefined \def \bsertitle#1{#1}\fi
\ifx \bsnm \undefined \def \bsnm#1{#1}\fi
\ifx \bsuffix \undefined \def \bsuffix#1{#1}\fi
\ifx \bparticle \undefined \def \bparticle#1{#1}\fi
\ifx \barticle \undefined \def \barticle#1{#1}\fi
\bibcommenthead
\ifx \bconfdate \undefined \def \bconfdate #1{#1}\fi
\ifx \botherref \undefined \def \botherref #1{#1}\fi
\ifx \url \undefined \def \url#1{\textsf{#1}}\fi
\ifx \bchapter \undefined \def \bchapter#1{#1}\fi
\ifx \bbook \undefined \def \bbook#1{#1}\fi
\ifx \bcomment \undefined \def \bcomment#1{#1}\fi
\ifx \oauthor \undefined \def \oauthor#1{#1}\fi
\ifx \citeauthoryear \undefined \def \citeauthoryear#1{#1}\fi
\ifx \endbibitem  \undefined \def \endbibitem {}\fi
\ifx \bconflocation  \undefined \def \bconflocation#1{#1}\fi
\ifx \arxivurl  \undefined \def \arxivurl#1{\textsf{#1}}\fi
\csname PreBibitemsHook\endcsname

\bibitem[\protect\citeauthoryear{Sun et~al.}{2022}]{Sun_Updating_2022}
\begin{barticle}
\bauthor{\bsnm{Sun}, \binits{X.-Y.}},
\bauthor{\bsnm{Ma}, \binits{R.-L.}},
\bauthor{\bsnm{He}, \binits{J.}},
\bauthor{\bsnm{Ding}, \binits{Y.-S.}},
\bauthor{\bsnm{Rui}, \binits{D.-S.}},
\bauthor{\bsnm{Li}, \binits{Y.}},
\bauthor{\bsnm{Yan}, \binits{Y.-Z.}},
\bauthor{\bsnm{Mao}, \binits{Y.-D.}},
\bauthor{\bsnm{Liao}, \binits{S.-Y.}},
\bauthor{\bsnm{He}, \binits{X.}},
\bauthor{\bsnm{Guo}, \binits{S.-X.}},
\bauthor{\bsnm{Guo}, \binits{H.}}:
\batitle{Updating {Framingham} {CVD} risk score using waist circumference and
  estimated cardiopulmonary function: a cohort study based on a southern
  {Xinjiang} population}.
\bjtitle{BMC Public Health}
\bvolume{22}(\bissue{1}),
\bfpage{1715}
(\byear{2022})
\doiurl{10.1186/s12889-022-14110-y} .
Accessed 2025-03-03
\end{barticle}
\endbibitem

\bibitem[\protect\citeauthoryear{Garcha and
  Phillips}{2023}]{Garcha_Social_2023}
\begin{barticle}
\bauthor{\bsnm{Garcha}, \binits{I.}},
\bauthor{\bsnm{Phillips}, \binits{S.}}:
\batitle{Social bias in artificial intelligence algorithms designed to improve
  cardiovascular risk assessment relative to the framingham risk score: a
  protocol for a systematic review}.
\bjtitle{BMJ Open}
\bvolume{13},
\bfpage{067638}
(\byear{2023})
\doiurl{10.1136/bmjopen-2022-067638}
\end{barticle}
\endbibitem

\bibitem[\protect\citeauthoryear{Khan et~al.}{2024}]{Khan_Heart_2024}
\begin{barticle}
\bauthor{\bsnm{Khan}, \binits{H.}},
\bauthor{\bsnm{Javaid}, \binits{N.}},
\bauthor{\bsnm{Bashir}, \binits{T.}},
\bauthor{\bsnm{Akbar}, \binits{M.}},
\bauthor{\bsnm{Alrajeh}, \binits{N.}},
\bauthor{\bsnm{Aslam}, \binits{S.}}:
\batitle{Heart disease prediction using novel ensemble and blending based
  cardiovascular disease detection networks: Enscvdd-net and blcvdd-net}.
\bjtitle{IEEE Access}
\bvolume{12},
\bfpage{109230}--\blpage{109254}
(\byear{2024})
\doiurl{10.1109/ACCESS.2024.3421241}
\end{barticle}
\endbibitem

\bibitem[\protect\citeauthoryear{Skouteli
  et~al.}{2024}]{Skouteli_Explainable_2024}
\begin{barticle}
\bauthor{\bsnm{Skouteli}, \binits{C.}},
\bauthor{\bsnm{Prentzas}, \binits{N.}},
\bauthor{\bsnm{Kakas}, \binits{A.}},
\bauthor{\bsnm{Pattichis}, \binits{C.}}:
\batitle{Explainable ai modeling in the prediction of cardiovascular disease
  risk}.
\bjtitle{Studies in Health Technology and Informatics}
\bvolume{316},
\bfpage{978}--\blpage{982}
(\byear{2024})
\end{barticle}
\endbibitem

\bibitem[\protect\citeauthoryear{Akkur}{2023}]{Akkur_Prediction_2023}
\begin{botherref}
\oauthor{\bsnm{Akkur}, \binits{E.}}:
Prediction of cardiovascular disease based on voting ensemble model and shap
  analysis.
Sakarya University Journal of Computer and Information Sciences
\textbf{6}
(2023)
\doiurl{10.35377/saucis...1367326}
\end{botherref}
\endbibitem

\bibitem[\protect\citeauthoryear{Chen et~al.}{2024}]{Chen_Self-supervised_2024}
\begin{barticle}
\bauthor{\bsnm{Chen}, \binits{L.-C.}},
\bauthor{\bsnm{Hung}, \binits{K.-H.}},
\bauthor{\bsnm{Tseng}, \binits{Y.-J.}},
\bauthor{\bsnm{Wang}, \binits{H.-Y.}},
\bauthor{\bsnm{Lu}, \binits{T.-M.}},
\bauthor{\bsnm{Huang}, \binits{W.-C.}},
\bauthor{\bsnm{Tsao}, \binits{Y.}}:
\batitle{Self-supervised learning-based general laboratory progress pretrained
  model for cardiovascular event detection}.
\bjtitle{IEEE Journal of Translational Engineering in Health and Medicine}
\bvolume{12},
\bfpage{43}--\blpage{55}
(\byear{2024})
\doiurl{10.1109/JTEHM.2023.3307794}
\end{barticle}
\endbibitem

\bibitem[\protect\citeauthoryear{Cai et~al.}{2024}]{Cai_Artificial_2024}
\begin{barticle}
\bauthor{\bsnm{Cai}, \binits{Y.}},
\bauthor{\bsnm{Cai}, \binits{Y.-Q.}},
\bauthor{\bsnm{Tang}, \binits{L.-Y.}},
\bauthor{\bsnm{Wang}, \binits{Y.-H.}},
\bauthor{\bsnm{Gong}, \binits{M.}},
\bauthor{\bsnm{Jing}, \binits{T.-C.}},
\bauthor{\bsnm{Li}, \binits{H.-J.}},
\bauthor{\bsnm{Li-Ling}, \binits{J.}},
\bauthor{\bsnm{Hu}, \binits{W.}},
\bauthor{\bsnm{Yin}, \binits{Z.}},
\bauthor{\bsnm{Gong}, \binits{D.-X.}},
\bauthor{\bsnm{Zhang}, \binits{G.-W.}}:
\batitle{Artificial intelligence in the risk prediction models of
  cardiovascular disease and development of an independent validation screening
  tool: a systematic review}.
\bjtitle{BMC Medicine}
\bvolume{22}(\bissue{1}),
\bfpage{56}
(\byear{2024})
\doiurl{10.1186/s12916-024-03273-7} .
Accessed 2025-03-02
\end{barticle}
\endbibitem

\bibitem[\protect\citeauthoryear{Kasim et~al.}{2023}]{Kasim_Validation_2023}
\begin{barticle}
\bauthor{\bsnm{Kasim}, \binits{S.}},
\bauthor{\bsnm{Ibrahim}, \binits{N.}},
\bauthor{\bsnm{Malek}, \binits{S.}},
\bauthor{\bsnm{Ibrahim}, \binits{K.}},
\bauthor{\bsnm{Aziz}, \binits{F.}},
\bauthor{\bsnm{Song}, \binits{C.}},
\bauthor{\bsnm{Chia}, \binits{Y.C.}},
\bauthor{\bsnm{Ramli}, \binits{A.}},
\bauthor{\bsnm{Negishi}, \binits{K.}},
\bauthor{\bsnm{Nasir}, \binits{N.}}:
\batitle{Validation of the general framingham risk score (frs), score2, revised
  pce and who cvd risk scores in an asian population}.
\bjtitle{The Lancet Regional Health - Western Pacific}
\bvolume{35},
\bfpage{100742}
(\byear{2023})
\doiurl{10.1016/j.lanwpc.2023.100742}
\end{barticle}
\endbibitem

\bibitem[\protect\citeauthoryear{Vaghefi
  et~al.}{2024}]{Vaghefi_Development_2024}
\begin{barticle}
\bauthor{\bsnm{Vaghefi}, \binits{E.}},
\bauthor{\bsnm{Squirrell}, \binits{D.}},
\bauthor{\bsnm{Yang}, \binits{S.}},
\bauthor{\bsnm{An}, \binits{S.}},
\bauthor{\bsnm{Xie}, \binits{L.}},
\bauthor{\bsnm{Durbin}, \binits{M.K.}},
\bauthor{\bsnm{Hou}, \binits{H.}},
\bauthor{\bsnm{Marshall}, \binits{J.}},
\bauthor{\bsnm{Shreibati}, \binits{J.}},
\bauthor{\bsnm{McConnell}, \binits{M.V.}},
\bauthor{\bsnm{Budoff}, \binits{M.}}:
\batitle{Development and validation of a deep-learning model to predict 10-year
  atherosclerotic cardiovascular disease risk from retinal images using the uk
  biobank and eyepacs 10k datasets}.
\bjtitle{Cardiovascular Digital Health Journal}
\bvolume{5}(\bissue{2}),
\bfpage{59}--\blpage{69}
(\byear{2024})
\doiurl{10.1016/j.cvdhj.2023.12.004}
\end{barticle}
\endbibitem

\bibitem[\protect\citeauthoryear{Duyar et~al.}{2024}]{Duyar_Detection_2024}
\begin{barticle}
\bauthor{\bsnm{Duyar}, \binits{C.}},
\bauthor{\bsnm{Senica}, \binits{S.}},
\bauthor{\bsnm{Kalkan}, \binits{H.}}:
\batitle{Detection of cardiovascular disease using explainable artificial
  intelligence and gut microbiota data}.
\bjtitle{Intelligence-Based Medicine}
\bvolume{10},
\bfpage{100180}
(\byear{2024})
\doiurl{10.1016/j.ibmed.2024.100180}
\end{barticle}
\endbibitem

\bibitem[\protect\citeauthoryear{Zhou et~al.}{2023}]{Zhou_Semi-supervised_2023}
\begin{botherref}
\oauthor{\bsnm{Zhou}, \binits{R.}},
\oauthor{\bsnm{Lu}, \binits{L.}},
\oauthor{\bsnm{Liu}, \binits{Z.}},
\oauthor{\bsnm{Xiang}, \binits{T.}},
\oauthor{\bsnm{Liang}, \binits{Z.}},
\oauthor{\bsnm{Clifton}, \binits{D.A.}},
\oauthor{\bsnm{Dong}, \binits{Y.}},
\oauthor{\bsnm{Zhang}, \binits{Y.-T.}}:
Semi-Supervised Learning for Multi-Label Cardiovascular Diseases Prediction:A
  Multi-Dataset Study
(2023).
\url{https://arxiv.org/abs/2306.10494}
\end{botherref}
\endbibitem

\bibitem[\protect\citeauthoryear{Hughes et~al.}{2025}]{Hughes_comparative_2025}
\begin{barticle}
\bauthor{\bsnm{Hughes}, \binits{C.M.L.}},
\bauthor{\bsnm{Zhang}, \binits{Y.}},
\bauthor{\bsnm{Pourhossein}, \binits{A.}},
\bauthor{\bsnm{Jurasova}, \binits{T.}}:
\batitle{A comparative analysis of binary and multi-class classification
  machine learning algorithms to detect current frailty status using the
  english longitudinal study of ageing (elsa)}.
\bjtitle{Frontiers in Aging}
\bvolume{6},
\bfpage{1501168}
(\byear{2025})
\doiurl{10.3389/fragi.2025.1501168}
\end{barticle}
\endbibitem

\bibitem[\protect\citeauthoryear{Fatima et~al.}{2023}]{Fatima_XGboost_2023}
\begin{barticle}
\bauthor{\bsnm{Fatima}, \binits{S.}},
\bauthor{\bsnm{Hussain}, \binits{A.}},
\bauthor{\bsnm{Amir}, \binits{S.}},
\bauthor{\bsnm{Ahmed}, \binits{S.H.}},
\bauthor{\bsnm{Aslam}, \binits{S.}}:
\batitle{Xgboost and random forest algorithms: An in depth analysis}.
\bjtitle{Pakistan Journal of Scientific Research}
\bvolume{3},
\bfpage{26}--\blpage{31}
(\byear{2023})
\doiurl{10.57041/pjosr.v3i1.946}
\end{barticle}
\endbibitem

\bibitem[\protect\citeauthoryear{Ar{\'e}valo-Cordovilla and
  Pe{\~n}a}{2025}]{Arevalo_evaluating_2025}
\begin{barticle}
\bauthor{\bsnm{Ar{\'e}valo-Cordovilla}, \binits{F.E.}},
\bauthor{\bsnm{Pe{\~n}a}, \binits{M.}}:
\batitle{Evaluating ensemble models for fair and interpretable prediction in
  higher education using multimodal data}.
\bjtitle{Scientific Reports}
\bvolume{15}(\bissue{1}),
\bfpage{29420}
(\byear{2025})
\doiurl{10.1038/s41598-025-15388-9}
\end{barticle}
\endbibitem

\end{thebibliography}
\end{document}